\documentclass{article}

     \usepackage[preprint]{neurips_2020}

\usepackage[utf8]{inputenc} 
\usepackage[T1]{fontenc}    
\usepackage{hyperref}       
\usepackage{url}            
\usepackage{booktabs}       
\usepackage{amsfonts}       
\usepackage{nicefrac}       
\usepackage{microtype}      
\usepackage{graphicx}
\usepackage{amsbsy}
\usepackage{amsmath}

\title{A Triangular Network For Density Estimation}

\author{%
  Xi-Lin Li  \\
  \texttt{lixilinx@gmail.com} \\
}

\begin{document}

\maketitle

\begin{abstract}
We report a triangular neural network implementation of neural autoregressive flow (NAF). Unlike many universal autoregressive density models, our design is highly modular, parameter economy, computationally efficient, and applicable to density estimation of data with high dimensions. It achieves state-of-the-art  bits-per-dimension indices on MNIST and CIFAR-10 (about $1.1$ and $3.7$, respectively) in the category of general-purpose density estimators.
\end{abstract}

\section{Introduction}

Probability density distribution (pdf) and conditional pdf estimations are topics of general interests, and also lie at the heart of many statistical learning and inference tasks. Traditional density estimation methods include histogram, kernel density estimation, orthogonal series density estimators, and finite mixture models (FMM) [1, 2, 3, 4]. These methods work well for data with low dimensions, but generally have difficulties when applied to data of high dimensions, i.e., the curse of dimensionality. Neural networks have achieve successes in many fields, and a few early examples of applying neural network to density estimations are independent component analysis (ICA) [5], Gaussianization [6] and [18]. Recently, there is a surge of neural network based density estimators. Many general-purpose neural network density estimators are based on autoregressive models, typically combining with normalizing flow, to name a few, neural networks for autoregressive density estimation (NADE) [18], masked autoencoder for distribution estimation (MADE) [7], masked autoregressive flow (MAF) for density estimation [8], neural autoregressive flows (NAF) [9], block neural autoregressive flow (B-NAF) [10], and unconstrained monotonic neural network (UMNN) [11]. Neural network density estimators can be designed for data with specific formats as well, e.g., non-linear independent components estimation (NICE) [12] and its successor, real NVP (Real NVP) [13], for images, where the correlations among neighboring pixels are hard-coded into the models. The neural ordinary differential equation (NODE) [14, 17] and its computationally cheaper version, free-form Jacobian of reversible dynamics (FFJORD) [16], provide yet another optimal transport based method for density estimation. These models and UMNN are different from traditional neural networks in the sense that they rely on numerical methods in both the train and inference phases. It is possible to combine several heterogeneous models to have hybrid, potentially more powerful, density estimators, e.g., the transformation autoregressive networks (TAN) [15]. Lastly, we recommend [21] for an overview of the more general normalizing flow based density estimation methods.          
 
NAF looks like a promising direction for general-purpose density estimation. However, its previous implementations are never applied to high dimensional data [9, 10], possibly due to their high memory footprints. In this paper, we start from a simple two layer autoregressive network unit, and show that it is a universal approximator for invertible transformations with triangular Jacobians. Then, deep neural networks constructed by properly cascading such triangular network units and permutations are proposed as universal density estimators. Our method is closely related to NAF [9] and B-NAF [10]. But, we adhere to the simplest autoregressive network form to keep the model memory footprint to the minimum. This is essential for high dimensional density estimations.  

\section{Background}

Here, we repeat a few well known facts to make our paper complete. This also serves to introduce our notations. 

\subsection{Probability density functions and change of variables}

We consider two continuous random variable $\pmb x$ and $\pmb y$ connected by a smooth invertible mapping $\pmb f(\cdot; \pmb \theta)$ with parameter vector $\pmb \theta$ as 
\begin{equation}
\pmb y = \pmb f(\pmb x; \pmb \theta)
\end{equation}
The probability density functions of $\pmb x$ and $\pmb y$, $p_X(\cdot)$ and $p_Y(\cdot)$ respectively, are related by
\begin{equation}\label{pdf_change_var}
 p_X(\pmb x) = p_Y(\pmb y) \left| \det \left( \frac{\partial \pmb y}{\partial \pmb x} \right) \right|
\end{equation}
where $\pmb \nabla \pmb f=\frac{\partial \pmb y}{\partial \pmb x}$ is called the Jacobian matrix, and $|\det(\cdot)|$ takes the absolute value of the determinant of a square matrix. Note that (\ref{pdf_change_var}) is  true only for invertible mappings. 

Relationship (\ref{pdf_change_var}), also referred to as normalizing flow in some work, is useful for density estimation and data generation. Assume that $p_X(\cdot)$ is unknown, but samples drawn from it are given. For $p_Y(\cdot)$ and $\pmb f(\cdot; \pmb \theta)$ with proper forms, we can approximate $p_X(\cdot)$ by maximizing the empirical likelihood given by the right side of (\ref{pdf_change_var}) with respect to $\pmb\theta$ on samples drawn from $p_X(\cdot)$. On the other hand, the inverse mapping of $\pmb f(\cdot; \pmb \theta)$, i.e., $\pmb x = \pmb f^{-1}(\pmb y; \pmb \theta)$, provides a mean for sampling from $p_X(\cdot)$ indirectly. It is common to model $\pmb f(\cdot; \pmb \theta)$ as neural networks due to their expressiveness.    

\subsection{Autoregressive  model for density estimation}  

Assume that $\pmb x = [x_1, x_2, \ldots, x_N]\in \mathbb{R}^N$, where $N$ is a positive integer. It is always feasible to factorize $p_X(\pmb x)$ recursively as 
\begin{equation}\label{auto_mdl}
 p_{X,1:n}(x_1, \ldots, x_n) = p_{X,1:n-1}(x_1, \ldots, x_{n-1}) p_{X,n|1:n-1}(x_n|x_1, \ldots, x_{n-1})
\end{equation}
Correspondingly, it is possible to map each vector $(x_1, \ldots, x_n)$ to $(y_1, \cdots, y_n)$ successively for all $1\le n\le N$ such that $p_Y(\pmb y)$ is factorized in the same fashion. Needless to say, all these $N$ mappings should be invertible. Together, they define an autoregressive  model. Noting that $\frac{\partial y_i}{\partial x_j}=0$ for all $1\le i<j\le N$, the Jacobian is a lower triangular matrix. This facilitates the calculation of its determinant, i.e.,
\begin{equation}
	\det \left( \frac{\partial \pmb y}{\partial \pmb x} \right) = \prod_{n=1}^N \frac{\partial y_n}{\partial x_n}
\end{equation} 
Otherwise, calculating the determinant of an arbitrary dense matrix has complexity $\mathcal{O}(N^3)$. Hence, autoregressive  model is one of the preferred choices for general-purpose density estimations with large $N$. 

\section{Monotonic triangular network}

\subsection{Monotonic triangular network unit}

We consider a special two layer feedforward neural network defined as
\begin{equation}\label{mono_net}
 \pmb y = \pmb V \pmb\phi \left( \pmb U \pmb x + \pmb a \right) + \pmb b
\end{equation}
where both random variables $\pmb x$ and $\pmb y$ have the same dimension $N$, $\pmb U$ and $\pmb V$ are two lower triangular block matrices with block sizes $(B, 1)$ and $(1, B)$, respectively, $B$ is a positive integer, and set $\{ \pmb U, \pmb V, \pmb a, \pmb b \}$ collects all the trainable parameters. Clearly, Jacobian $\frac{\partial \pmb y}{\partial \pmb x}$ is a lower triangular matrix as well.  Let the $n$th diagonal blocks  of $\pmb U$ and $\pmb V$ be $[u_{n,1}, u_{n,2}, \ldots, u_{n,B}]$ and $[v_{n,1}, v_{n,2}, \ldots, v_{n,B}]$, respectively. Then, we have
\begin{equation}\label{shift_scale_dtanh}
 \frac{\partial y_n}{\partial x_n} = \sum_{i=1}^B u_{n,i} v_{n,i} \dot{\phi}_{(n-1)B+i}, \quad 1\le n\le N
\end{equation}
where $\dot{\phi}_{(n-1)B+i}$ is the derivative of the $[(n-1)B+i]$th nonlinearity (activation function) in $\pmb\phi$. To simplify the notations, we do not explicitly show $\dot{\phi}_{(n-1)B+i}$'s dependence on $\pmb U$, $ \pmb x$ and $\pmb a $. An interesting observation is that when $\phi_{(n-1)B+i}$ is monotonically increasing, and $u_{n,i}$ and $v_{n,i}$ have the same or opposite sign for all $1\le i\le B$, $\frac{\partial y_n}{\partial x_n}$ will always be either positive or negative. Then, $y_n$ must be monotonic with respect to $x_n$. This observation is true for all $1\le n\le N$. Hence, (\ref{mono_net}) defines a monotonic network when $\pmb\phi$ is element-wisely monotonically increasing, and each group of $u_{n,i}$ and $v_{n,i}$ for $1\le i\le B$ have the same or opposite sign. The following proposition states that such networks can approximate any invertible mappings with triangular Jacobians. The proof should be straightforward following the universal approximation theorem and the fact that networks with positive weights can approximate any continuous monotonic mappings [20]. Here, we sketch an informal one to make our report complete. 

\emph{Proposition 1:} The neural network defined in (\ref{mono_net}) can approximate any continuous invertible mapping on a compact subset of $\mathbb{R}^N$ with lower triangular Jacobians arbitrarily well when C1) $B$ is sufficiently large; C2) each group of $u_{n,i}$ and $v_{n,i}$, $1\le i\le B$, have the same or opposite sign for all $1\le n\le N$; C3) all the nonlinearities in $\pmb\phi$ are monotonically increasing, and lower and upper bounded.

\emph{Proof:} The universal approximation theorem ensures that without imposing the sign constraints on the block diagonals of $\pmb U$ and $\pmb V$, the network defined in (\ref{mono_net}) can approximate any continuous mapping with triangular Jacobian. Hence, we only need to show that $y_n$ can approximate any mapping monotonic with respect to $x_n$, or equivalently, $\frac{\partial y_n}{\partial x_n}$ can approximate any scalar function of $(x_1, \ldots, x_n)$ whose outputs always have the same sign. Let us rewrite $y_n$ and $\frac{\partial y_n}{\partial x_n}$ explicitly as below
\begin{eqnarray} 
y_n & = & \sum_{i=1}^B v_{n,i} \phi_{(n-1)B+i}\left( u_{n,i} ( x_n - c_i(x_1, \ldots, x_{n-1}) ) \right) + ({\rm a\;   bias\; constant}) \\ \label{dyn_dxn}
\frac{\partial y_n}{\partial x_n} & = & \sum_{i=1}^B u_{n,i} v_{n,i} \dot{\phi}_{(n-1)B+i}\left( u_{n,i} ( x_n - c_i(x_1, \ldots, x_{n-1}) ) \right)
\end{eqnarray}
where $c_i$, $1\le i\le B$, are $B$ scalar functions of $(x_1, \ldots, x_{n-1})$ introduced after reparameterization. Without loss of generality, we assume that for all $1\le i\le B$, ${\phi}_{(n-1)B+i}$ is  bounded as $\phi_{(n-1)B+i}(-\infty)=0$ and $\phi_{(n-1)B+i}(\infty)=1$, and $u_{n,i}$ and $v_{n, i}$ are positive, i.e., $y_n$ is monotonically increasing with respect to $x_n$. Then, for sufficiently large $u_{n,i}$, we have
\begin{equation}\label{Dirac}
 \lim_{u_{n,i}\rightarrow \infty}u_{n,i}  \dot{\phi}_{(n-1)B+i}\left[ u_{n,i} ( x_n - c_i(x_1, \ldots, x_{n-1}) ) \right] = \delta(x_n - c_i(x_1, \ldots, x_{n-1}) )
\end{equation}
where $\delta(\cdot)$ is the Dirac delta function. Now, let us consider an arbitrary continuous scalar mapping $(x_1, x_2, \ldots, x_n) \mapsto g(x_1, x_2, \ldots, x_n)>0$. It is clear that any value $g(z_1, z_2, \ldots, z_n)$ can be approximated by letting $u_{n,i}v_{n, i} = g(z_1, z_2, \ldots, z_n)$ and $c_i(x_1, \ldots, x_{n-1}) $ be a function that equals $z_n$ only when $(x_1, \ldots, x_{n-1})=(z_1, \ldots, z_{n-1})$, which is feasible due to the universal approximation theorem. Thus, (\ref{dyn_dxn}) is dense on a compact set when $B$ is sufficiently large. This finishes the proof. 
$\square$

Let us check each condition used in the above proof closely. C1) is quite standard. C2) arises naturally as outputs of the target mapping have the same sign. C3) is used in (\ref{Dirac}). Intuitively, we see that the right side of (\ref{shift_scale_dtanh}) is a weighted sum of the shifted and rescaled versions of the derivative of nonlinearity ${\phi}_{(n-1)B+i}$. Under mild conditions, these shifted and rescaled derivatives  form an over complete base for function approximation, as illustrated in Fig.~1 for the case with dimension 1. However, unbounded nonlinearities are widely used in today's neural networks, e.g., the rectified linear unit (ReLU). Unfortunately, it is not difficult to show that such nonlinearities do not work here. The derivative of ReLU is monotonic. Hence, derivative of its weighted sum is monotonic as well as long as the weights have the same sign. But, not all monotonic functions have monotonic derivatives. Thus, any nonlinearities with monotonic derivatives cannot be used here.  

\begin{figure}
	\centering
	\includegraphics[width=0.5\textwidth]{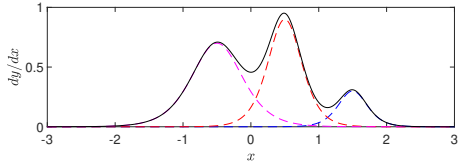}
	\caption{A positive scalar function (solid line) is approximated as the sum of three shifted and rescaled versions of the derivative of $\tanh$ (dotted lines).}
\end{figure}

\subsection{Unconstrained monotonic network for density estimation}

For density estimation, it is sufficient to constrain $u_{n, i}$ and $v_{n, i}$ to be positive for all $1\le n\le N$ and $1\le i\le B$. It could be inconvenient to keep all these positiveness constraints on the block diagonals of $\pmb U$ and $\pmb V$. But, it is trivial to reparameterize them to drop off these constraints. In our implementations, we choose reparameterization
\begin{equation}
 u_{n, i} = -\log \sigma(-\mu_{n,i}), \quad  v_{n, i} = -\log \sigma(-\nu_{n,i}), \quad 1\le n\le N, \;  1\le i\le B
\end{equation}  
where $\sigma(\cdot)$ is the sigmoid function, and $\mu_{n,i}$ and $\nu_{n,i}$ are free to take any real values. Hyperparameter $B$ can be viewed as the order of our monotonic triangular network unit. Note that memory footprint of a unit is proportional to $B$. Regarding $\pmb\phi$, commonly used squash functions, e.g., the sigmoid and tanh,  will work since they are monotonic and bounded. Certain unbounded nonlinearities, e,g., ReLU, should be avoided per our above discussions.  

It is a common practice to choose $p_Y(\pmb y)$ as a standard normal density. Then, given samples $\pmb x$ from an unknown distribution $p_X(\cdot)$, we can estimate it in the form of (\ref{pdf_change_var}) by minimizing expectation of the following  negative-logarithm-likelihood (NLL)
\begin{equation}\label{gauss_nll}
  -\sum_{n=1}^N \log \frac{\partial y_n}{\partial x_n} + 0.5\pmb y^T \pmb y + 0.5N \log(2\pi)
\end{equation}  
with respect to parameters $\{ \pmb U, \pmb V, \pmb a, \pmb b \}$. On the other hand, a monotonic network unit can be reversed in a way similar to backward substitution for solving linear triangular  equations, i.e., solving for $x_n$ when $(x_1, \ldots, x_{n-1})$ are ready. As $y_n$ is monotonic with respect to $x_n$, simple root finding methods like bisection could be sufficient for solving for $x_n$ given $y_n$ and $(x_1, \ldots, x_{n-1})$. 

Lastly, deep neural networks obtained by stacking several monotonic triangular network units are still monotonic, have triangular Jacobians, and can be inverted with a backward substitution like routine as well. Such networks might look superficially similar to the B-NAF in [10]. But, there are a few subtle differences. We do not apply any nonlinear mapping on the outputs of each monotonic triangular network unit. Intuitively, squashing will deviate outputs of a unit away from normal distribution, which conflicts with our aim. Memory footprint of each unit is proportional to its order $B$. This avoids the memory explosion issue caused by those wide intermediate layers in B-NAF. A few implementation tips proposed in the next section further enhance the usefulness of this design for high dimensional density estimation.         

\section{Practical triangular networks for density estimation and data generation}

Here, we prescribe a few modifications to the monotonic triangular networks in Section 3 to enhance its usefulness in practice.   

\subsection{Non-autoregressive model by inserting permutations}

Although a large enough monotonic triangular network suffices to universal density estimation, it is unlikely the case that autoregressive model is parameter economy for all density estimation tasks. In general, we can obtain deep reversible non-autoregressive models by cascading monotonic triangular network units and permutations as below
\[ \xrightarrow{\pmb x_0}  \fbox{MonoTriNetU}\rightarrow \fbox{P} \xrightarrow{\pmb x_1} \fbox{MonoTriNetU}\rightarrow \fbox{P} \xrightarrow{\pmb x_2} \ldots \xrightarrow{\pmb x_{L-1}} \fbox{MonoTriNetU} \rightarrow \fbox{P} \xrightarrow{\pmb x_L} \]
where each \fbox{MonoTriNetU} and \fbox{P} denote an independent monotonic triangular network unit and permutation unit, respectively. Due to these inserted permutations, the Jacobian $\frac{\partial \pmb x_L}{\partial \pmb x_0}$ is no longer ensured to be triangular. However, it is still convenient to calculate its determinant as
\begin{equation}
 \det\left(\frac{\partial \pmb x_L}{\partial \pmb x_0}\right) = \prod_{\ell=1}^{L} \det\left(\frac{\partial \pmb x_\ell}{\partial \pmb x_{\ell - 1}}\right) = \prod_{\ell=1}^{L} \prod_{n=1}^{N} \frac{\partial x_{\ell, n}}{\partial x_{\ell-1, n}}
\end{equation}     
where we have used the fact that the determinant of a permutation matrix is $1$. Different from a monotonic triangular network, a one-pass backward substitution like inversion no longer works here. Instead, we need to inverse each permutation and monotonic triangular network unit step by step, down from the top layer to the bottom layer.  

Let us check the network 
\begin{equation}\label{flip_mono_net}
\xrightarrow{\pmb x}  \fbox{MonoTriNetU}\rightarrow \fbox{Flip} \rightarrow \fbox{MonoTriNetU}\rightarrow \fbox{Flip} \xrightarrow{\pmb y} 
\end{equation}
to see why permutation could help, where \fbox{Flip} is the permutation that reverses the order of a vector. Mathematically, we have $\pmb y=\pmb F\pmb f_2\left( \pmb F \pmb f_1(\pmb x)\right)$, where $\pmb F={\rm antidiag}(1, 1, \ldots, 1)$ is the unit anti-diagonal matrix, and $\pmb f_1(\cdot)$ and $\pmb f_2(\cdot)$ are two independent monotonic triangular network units. Its Jacobian is 
\begin{equation}\label{ul}
\frac{\partial \pmb y}{\partial \pmb x} = \left(\pmb F \pmb\nabla\pmb f_2 \pmb F\right) \pmb \nabla \pmb f_1
\end{equation}
where $\pmb \nabla \pmb f_1$ is a lower triangular matrix, and $\pmb F \pmb\nabla\pmb f_2 \pmb F$ is an upper triangular one. We know that a constant square matrix can be factorized as the product of an upper triangular matrix and a lower  one\footnote{The product sometimes includes a permutation matrix as well.}. Similarly, we expect that the upper-lower decomposition form in (\ref{ul}) can approximate the Jacobians from a larger family of invertible mappings. 

\subsection{Bijective network with a `log' nonlinearity for data generation}

It is worthy to point out that with a finite $B$ and bounded nonlinearities, (\ref{mono_net}) only defines an injective mapping on $\mathbb{R}^N$. Specifically, with nonlinearity `tanh', we have $|y_n|< \sum_{i=1}^B |v_{n,i}|$. Thus, when the target distribution is a standard normal one, it is not possible to invert all samples drawn from $p_Y(\cdot)$. Although the probability measure of such non-invertible samples could be negligibly small for good enough density models, this brings inconveniences for applications like data generation. Here, we propose to replace a bounded nonlinearity with a `log' nonlinearity, ${\rm sign}(x)\log(1 + |x|)$, when bijection is a must. This `log' nonlinearity is neither lower nor upper bounded. It is easy to show that the networks in Section 4.1 become bijective on $\mathbb R^N$ after adopting it. Our experiences suggests that the `log' nonlinearity behaves similar to a squash one since its output amplitudes grow slowly enough with respect to its inputs. It even could outperform the tanh nonlinearity in negative-logarithm-likelihood performance index when $B$ is too small and $p_Y(\cdot)$ is the standard normal density. This is not astonishing since the sum of a few hard bounded terms cannot fit well with the tails of a normal distribution. 

\subsection{Compact model storage}

Two strategies  are used to save our neural networks compactly. First, instead of saving the two triangular matrices in (\ref{mono_net}) separately, we save most of their elements  compactly in a dense matrix. Let ${\rm off}(\pmb V) = \pmb V - {\rm bdiag}(\pmb V)$, where ${\rm bdiag}(\pmb V)$ is the block diagonals of $\pmb V$. Then, we can put ${\rm off}(\pmb V^T)$ in the upper triangular part of $\pmb U$ to save memory. The block diagonals of $\pmb V$ is saved as a separate vector. Second, we generate the masks for exacting $\pmb U$ and ${\rm off}(\pmb V^T)$ from the dense matrix they are nested in on the fly. Note that $B$, and thus the masks, can be inferred from the sizes of $\pmb U$ and $\pmb V$. Thus, there is no need to save them in advance. These two strategies reduce the model memory footprint to about one fourth of value required by naively saving $\pmb U$, $\pmb V$ and their masks in separated matrices. Clearly, it is the design regularity of our monotonic triangular network unit to make such significant storage reduction  possible. 

\subsection{Linear autoregressive input normalization}

Lastly, it is known that input normalization generally helps to facilitate  training. We propose to use a linear autoregressive model to  preprocess the input as $\pmb x\leftarrow \pmb\Gamma (\pmb x - \pmb m)$, where $\pmb m$ and $\pmb C$ are the (empirical) mean and covariance matrix of $\pmb x$, respectively, and $\pmb\Gamma$ is a lower triangular matrix   given by Cholesky decomposition $\pmb C^{-1} = \pmb\Gamma^{T} \pmb\Gamma$. Unlike the more popular principle component analysis (PCA) based whitening preprocessing, this one works seamlessly with our neural network since itself is a simple autoregressive unit. The preprocessing parameters, $\pmb m$ and $\pmb\Gamma$, have no need to be exact, nor need to be adapted during the training, since they could be absorbed into the input layer of the first monotonic triangular network unit as $\pmb U\leftarrow \pmb U\pmb \Gamma$ and $\pmb a\leftarrow \pmb a - \pmb U\pmb \Gamma\pmb m$. Indeed, They will not show up in the trained and finalized model coefficients after being absorbed.  
 
\section{Experimental Results}

Pytorch implementation of our method and more experimental results can be found at \url{https://github.com/lixilinx/TriNet4PdfEst}. Here, we report the density estimation results for  MNIST and CIFAR-10 data sets. The label information is discarded.  The pixel values in space $[0, 255]^N$ are first dequantized by adding uniform noise $\mathcal{U}(0, 1)$, then rescaled to range $[0, 1]$, and finally transferred to the logit space with mapping $ x\mapsto {\rm logit}(\lambda + (1-2\lambda)x)$, where $\lambda=10^{-6}$ and $0.05$ for MNIST and CIFAR-10, respectively. We estimate the density in the logit space, but report the bits-per-dimension performance index for comparison as it is more popular than the negative-logarithm-likelihood one. 

We have trained invertible triangular networks consists of four cascaded monotonic units with different settings. Block sizes are $100$ and $8$ for MNIST and CIFAR-10, respectively. Batch size is $64$ in both tasks. All models are trained with Adam starting from initial step size $10^{-4}$. We reduce the step size by one order of magnitude when no performance improvement is observed on the validation set. The train, validation and test sets are defined following the ways in [8]. We find that our model could seriously overfit the CIFAR-10 dataset even with early stopping and such a small $B$  (about $0.5$ bits-per-dimension gap between train and validation set). We have tried two ways to alleviate overfitting, L1 regularization and train data augmentation by randomly and circularly shifting the images up to $\lfloor 0.1\times ({\rm image\; size}) \rfloor$ pixels both horizontally and vertically. Detailed negative-logarithm-likelihood and bits-per-dimension indices on train, validation and test sets under different settings are summarized in  Appendix A. Table 1 summarizes the test results for comparison. Standard deviations of our method are not listed as all are below $0.01$. Note that only those general-purpose density estimators are compared here. Density estimators  specifically designed for images  could perform  better [21]. From Tabel 1, we see that our models outperform their competitors by great margins, especially on the CIFAR-10 task.   

Randomly generated handwritten digits and image patches drawn from the MNIST and CIFAR-10 density models trained with L1 regularization are posted in Appendix B. Models trained with augmented data are not used here since their performances on train set are worse than the ones with L1 regularization (see Appendix A). Most digit samples are recognizable, although less generated image patches make sense to a human subject.  

\begin{table}
	\caption{Test bits-per-dimension comparison on image density estimation benchmarks.  }
	\label{sample-table}
	\centering
	\begin{tabular}{lll}
		\toprule
		      & MNIST     & CIFAR-10  \\
		Best of MADE/MoG [8] & $1.41\pm 0.01$  & $5.67\pm 0.01$    \\
		Best of Real NVP [8] & $1.93\pm 0.01$ & $4.53 \pm 0.01$ \\
		Best of MAF/MoG [8] & $1.52\pm 0.01$ & $4.31\pm 0.01$ \\
		TAN [15] & $1.19$ & $3.98$ \\
		UMNN [11] & $1.13\pm 0.02$ & $-$ \\
		& & \\
		{\bf TriNet}, L1 regularization & $1.13 $ &  $3.70  $\\
		{\bf TriNet}, augmentation & ${\bf 1.09} $ &  ${\bf 3.69 } $\\
		\bottomrule
	\end{tabular}
\end{table}

\section*{References}

\medskip

\small

[1] B. W. Silverman, \emph{Density Estimation for Statistics and Data
	Analysis}. London: Chapman and Hall, 1986.

[2] J. E. Kolassa, \emph{Series Approximation Methods in Statistics},
3rd Edition. New York: Springer, 2006.

[3] A. J. Izenman, Recent developments in nonparametric density
estimation. {J. Amer. Statist. Assoc.}, vol. 86, pp.
205--225, 1991.

[4] G. J. McLachlan, \emph{Finite Mixture Models}, Wiley, 2000.

[5] H. Aapo and E. Oja. Independent component analysis: algorithms and applications. Neural Networks, no. 4--5, vol. 13, pp. 411--430, 2000. 

[6] S. S. Chen and R. A. Gopinath. Gaussianization. In Proceedings of NIPS, 2000.

[7] M. Germain, K. Gregor, I. Murray, and H. Larochelle. Made: masked autoencoder for distribution estimation. In International Conference on Machine Learning, pages 881--889, 2015. 

[8] G. Papamakarios, T. Pavlakou, and I. Murray. Masked autoregressive flow for density estimation. In Advances in Neural Information Processing Systems, pages 2338--2347, 2017. 

[9] C. W. Huang, D. Krueger, A. Lacoste, and A. Courville. Neural autoregressive flows. In International Conference on Machine Learning, pages 2083--2092, 2018. 

[10] N. De Cao, I. Titov, and W. Aziz. Block neural autoregressive flow. arXiv preprint, arXiv:1904.04676, 2019. 

[11] A. Wehenkel, and G. Louppe, Unconstrained monotonic neural networks. In Conference on Neural Information Processing Systems, Vancouver, Canada, 2019.

[12] L. Dinh, D. Krueger, and Y. Bengio. NICE: non-linear independent components estimation. In International Conference on Learning Representations (ICLR), 2015.

[13] L. Dinh, J. Sohl-Dickstein, and S. Bengio. Density estimation using Real NVP. In International Conference on Learning Representations, 2017.

[14] T. Q. Chen, Y. Rubanova, J. Bettencourt, and D. K. Duvenaud. Neural ordinary differential equations. In Advances in Neural Information Processing Systems, pages 6571--6583, 2018.

[15] J. Oliva, A. Dubey, M. Zaheer, B. Poczos, R. Salakhutdinov, E. Xing, and J. Schneider. Transformation autoregressive networks. In International Conference on Machine Learning, pp. 3895--3904, 2018. 

[16] W. Grathwohl, R. T. Chen, J. Bettencourt, I. Sutskever, and D. Duvenaud. FFJORD: free form continuous dynamics for scalable reversible generative models. In International Conference on Machine Learning, 2018. 

[17] L. Zhang, W. E, and L. Wang. Monge-Amp\`ere flow for generative modeling. arXiv preprint, arXiv:1809.10188, 2018.

[18] H. Larochelle, and I. Murray. The neural autoregressive distribution estimator. In Proceedings of the Fourteenth International Conference on Artificial Intelligence and Statistics, pp. 29--37, 2011. 

[19] M. M. Ismail and A. Atiya. Neural networks for density estimation. In Advances in Neural Information Processing Systems, pages 522--528, 1999.  

[20] H. Daniels, and M. Velikova, Monotone and partially monotone neural networks. IEEE Transactions on Neural Networks, no. 6, vol. 21, pp. 906--917, 2010. 

[21] I. Kobyzev, S. J. D. Prince, and M. A. Brubaker. Normalizing flows: an introduction and review of current methods. arXiv preprint,  arXiv:1908.09257v3, 2020.

\section*{Appendix A: further density estimation results } 

\subsection*{A1: Optional setups}

Nonlinearity (activation function): either tanh or log, i.e., ${\rm sign}(x)\log(1 + |x|)$.

Optional permutation: flipping  the outputs of each monotonic triangular network unit. 

L1 regularization term: $\eta\times $(sum of the absolute values of model coefficients), where $\eta> 0$.  L1 regularization encourages sparse connections, which imply sparse dependence structures in the data.      

Optional data augmentation: randomly and circularly shift the train images up to $\lfloor0.1({\rm image \;size})\rfloor$ pixels both horizontally and vertically.   

Note that the negative-logarithm-likelihood indices from different methods might not be directly comparable due to different processing. However, the bits-per-dimension indices are always comparable.  

\subsection*{A2: Results without regularization and augmentation}

We reduce the training step size when no validation performance gain is observed after  $10$ and $5$ successive epochs for MNIST and CIFAR-10, respectively. Flipping generally helps for CIFAR-10, but not for MNIST. It makes sense as most people write digits from the top left corner down to the bottom right  corner, comparable to an autoregressive generative model. The `log' nonlinearity significantly outperforms the tanh one on the CIFAR-10 task, perhaps due to a small $B$. Huge performance gaps between train and validation sets on the CIFAR-10 task suggest serious overfitting.   

\begin{table*}[h]
	\caption{Negative-logarithm-likelihood (bits-per-dimension), w/o regularization, w/o augmentation  	 }
	\centering
	\begin{tabular}{lllllll}
		\toprule
		& & MNIST & &     & CIFAR-10  & \\
		& Train & Validation & Test & Train & Validation & Test \\
		& & & & & & \\
	
		w/o flip, tanh  & $654(1.06)$ & $703(1.15)$ & $698(1.14)$ & $-4515(3.62)$ & $-3558(4.07)$ & $-3523(4.09)$ \\
		w/o flip, log  & $680(1.11)$&$709(1.16)$ &$705(1.16)$ & $-4584(3.59)$ & $-3719(3.99)$ & $-3744(3.98)$\\
		
		& & & & & & \\
			
		w flip, tanh & $706(1.16)$ & $721(1.18)$& $716(1.18)$& $-4490(3.63)$ & $-3401(4.14)$ & $-3453(4.12)$\\
		w flip, log  & $692(1.13)$ & $725(1.19)$ & $722(1.19)$ & $-4828(3.47)$ & $-3804(3.96)$ & $-3810(3.95)$ \\
		\bottomrule
	\end{tabular}
\end{table*}

\subsection*{A3: Results without regularization and with augmentation}

We reduce the training step size when no validation performance gain is observed after  $10$ and $5$ successive epochs for MNIST and CIFAR-10, respectively. Again, flipping helps for CIFAR-10, but not for MNIST. This data augmentation method seems quite coarse. Over fitting is avoid, but performances on the train sets drop a lot, compared with those from Table 2.   

\begin{table}[h]
	\caption{Negative-logarithm-likelihood (bits-per-dimension), w/o regularization, w augmentation  }
	\centering
	\begin{tabular}{lllllll}
		\toprule
		& & MNIST & &     & CIFAR-10  & \\
		& Train & Validation & Test & Train & Validation & Test \\
		& & & & & & \\
		
		w/o flip, tanh & $676(1.10)$ & $677(1.10)$ & ${\bf 674(1.10)}$& $-4150(3.79)$ & $-4135(3.80)$& $-4126(3.80)$ \\
		w/o flip, log & $674(1.10)$ & $675(1.10)$ & ${\bf 672(1.09)}$ & $-4166(3.78)$ & $-4184(3.77)$ & $-4152(3.79)$ \\
		
		& & & & & & \\
		
		w flip, tanh & $706(1.16)$ & $707(1.16)$ & $703(1.15)$ & $-4282(3.73)$ & $-4286(3.73)$ & $-4261(3.74)$ \\
		w flip, log & $721(1.18)$ & $722(1.19)$ & $717(1.18)$ & $-4387(3.68)$ & $-4391(3.68)$ & ${\bf -4365(3.69)}$ \\
		\bottomrule
	\end{tabular}
\end{table}

\subsection*{A4: Results with regularization and without augmentation}

We reduce the training step size when no validation performance gain is observed after  $10$ successive epochs. We try $\eta$ with possible values from set $\{\ldots, 10^{-3}, 10^{-4}, 10^{-5}, \ldots\}$, and choose the one performing the best on validation set. L1 regularization factors are $10^{-4}$ and $10^{-3}$ for MNIST and CIFAR-10, respectively.  It looks like that regularization improves the MNIST training performance when compared with Table 2. This is  caused by different early stopping conditions. Actually, the train performance indices on both tasks can go further down without early stopping. 

\begin{table*}[h]
	\caption{Negative-logarithm-likelihood (bits-per-dimension), w regularization, w/o augmentation  	 }
	\centering
	\begin{tabular}{lllllll}
		\toprule
		& & MNIST & &     & CIFAR-10  & \\
		& Train & Validation & Test & Train & Validation & Test \\
		& & & & & & \\
		
		w/o flip, tanh  & $647(1.05)$  & $694(1.13)$& ${\bf 690(1.13)}$ & & &  \\
		w/o flip, log  & & & & & & \\
		
		& & & & & & \\
		
		w flip, tanh & & & & & & \\
		w flip, log  & & & & $-4649(3.56)$ & $-4324(3.71)$ & ${\bf -4351(3.70)}$  \\
		\bottomrule
	\end{tabular}
\end{table*}

\section*{Appendix B: randomly generated image samples} 

\subsection*{B1: MNIST samples}

\begin{figure}[h]
	\centering
	\includegraphics[width=0.5\textwidth]{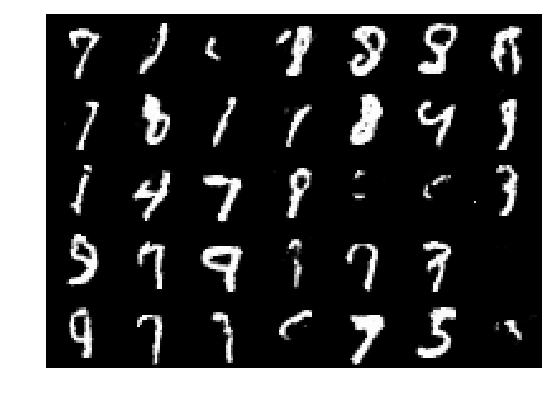}
	\caption{$35$ randomly generated handwritten digit images drawn from the MNIST density model trained with L1 regularization.  }
\end{figure}

\subsection*{B2: CIFAR-10 samples}

\begin{figure}[h]
	\centering
	\includegraphics[width=0.7\textwidth]{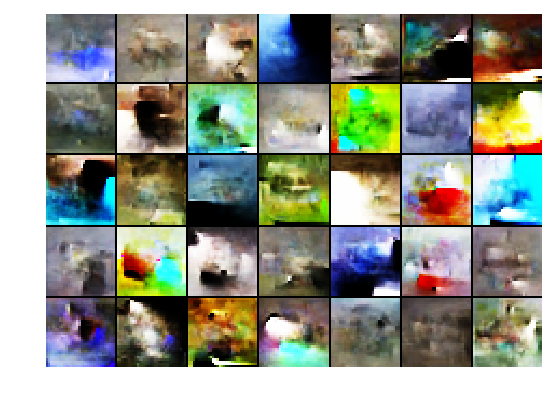}
	\caption{$35$ randomly generated image patches drawn from the CIFAR-10 density model trained with L1 regularization. A few look recognizable, e.g., the last one looks like a frog.   }
\end{figure}

\end{document}